\documentclass[sigconf,natbib=false]{acmart}

\setcopyright{acmcopyright}
\copyrightyear{2020}
\acmYear{2020}

\acmConference[CHIL '20 Workshop]{CHIL '20: ACM Conference on Health, Inference, and Learning; Workshop Track}{April 02--04, 2020}{Toronto, ON}
\acmBooktitle{CHIL '20: ACM Conference on Health, Inference, and Learning; Workshop Track. April 02--04, 2020, Toronto, ON}

\usepackage{booktabs}
\usepackage{lineno,hyperref}
\usepackage{multirow}
\usepackage{amsmath}

\usepackage[T1]{fontenc}

\usepackage{amsmath}

\usepackage{tgtermes}

\usepackage{scalefnt,letltxmacro}
\LetLtxMacro{\oldtextsc}{\textsc}
\renewcommand{\textsc}[1]{\oldtextsc{\scalefont{1.10}#1}}

\usepackage{listings}

\usepackage{subfiles}

\usepackage[english]{babel}
\usepackage[parfill]{parskip}
\usepackage{afterpage}
\usepackage{framed}
\usepackage{nicefrac}

\usepackage{lineno}

\usepackage{ragged2e}

\usepackage{graphicx}
\usepackage[labelfont=bf]{caption}
\usepackage[format=hang]{subcaption}
\usepackage{wrapfig}

\usepackage{booktabs}
\usepackage{array}

\usepackage{etoolbox,siunitx}
\robustify\bfseries
\usepackage[algoruled]{algorithm2e}
\usepackage{listings}

\usepackage{amsthm}

\addto\captionsenglish{\renewcommand*\proofname{proof}}  

\usepackage[all]{hypcap}
\hypersetup{citecolor=Violet}
\hypersetup{linkcolor=black}
\hypersetup{urlcolor=MidnightBlue}

\usepackage[capitalize,nameinlink]{cleveref}
\Crefname{prop}{Proposition}{Propositions}

\usepackage[acronym,nowarn]{glossaries}
\glsdisablehyper

\usepackage{listings}
\lstdefinestyle{alp_style}{
    commentstyle=\color{OliveGreen},
    numberstyle=\tiny\color{black!60},
    stringstyle=\color{BrickRed},
    basicstyle=\ttfamily\scriptsize,
    breakatwhitespace=false,
    breaklines=true,
    captionpos=b,
    keepspaces=true,
    numbers=none,
    numbersep=5pt,
    showspaces=false,
    showstringspaces=false,
    showtabs=false,
    tabsize=2
}
\usepackage{titlesec}
\titlespacing*{\section}
{0pt}{1ex plus 1ex minus .2ex}{1ex plus .2ex}
\titlespacing*{\subsection}
{0pt}{1ex plus 1ex minus .2ex}{1ex plus .2ex}
\newacronym{ELBO}{elbo}{evidence lower bound}
\newacronym{GMM}{gmm}{Gaussian mixture model}
\newacronym{KL}{kl}{Kullback-Leibler}
\newacronym{LDA}{lda}{latent Dirichlet allocation}
\newacronym{SVI}{svi}{stochastic variational inference}
\newacronym{DEF}{def}{deep exponential family}
\newacronym{rfs}{\textsc{rfs}}{\textsc{rankfromsets}}
\newacronym{ctpf}{\textsc{ctpf}}{collaborative topic Poisson factorization}
\newacronym{lightfm}{Light\textsc{fm}}{LightFM}
\newacronym{bpr}{\textsc{bpr}}{Bayesian Personalized Ranking}
\newacronym{lstm}{\textsc{lstm}}{long short-term memory}
\newacronym{bi-lstm}{\textsc{bi-lstm}}{bidirectional long short-term memory}
\newacronym{gru}{\textsc{gru}}{gated recurrent unit}
\newacronym{cb}{ClinicalB\textsc{ert}}{}
\newacronym{bert}{\textsc{bert}}{bidirectional encoder representations from transformers}
\newacronym{mimic}{\textsc{mimic-iii}}{Medical Information Mart for Intensive Care III}
\newacronym{ehr}{\textsc{ehr}}{electronic health record}

\usepackage[%
minnames=1,maxnames=99,maxcitenames=2,
style=numeric, 
doi=false,
url=false,
firstinits=true,
hyperref,
natbib,
backend=bibtex,
sorting=nyt
]{biblatex}%

\newbibmacro*{journal}{%
  \iffieldundef{journaltitle}
    {}
    {\printtext[journaltitle]{%
       \printfield[noformat]{journaltitle}%
       \setunit{\subtitlepunct}%
       \printfield[noformat]{journalsubtitle}}}}

\AtEveryBibitem{%
\ifentrytype{article}{
    \clearfield{url}%
    \clearfield{urldate}%
    \clearfield{eprint}
}{}
\ifentrytype{book}{
    \clearfield{url}%
    \clearfield{urldate}%
}{}
\ifentrytype{collection}{
    \clearfield{url}%
    \clearfield{urldate}%
}{}
\ifentrytype{incollection}{
    \clearfield{url}%
    \clearfield{urldate}%
}{}
}

\AtEveryBibitem{
    \clearfield{pages}
    \clearfield{review}%
    \clearfield{series}
    \clearfield{volume}
    \clearfield{pages}
    \clearfield{month}
    \clearfield{eprint}
    \clearfield{isbn}
    \clearfield{issn}
    \clearlist{location}
    \clearfield{series}
    \clearlist{publisher}
    \clearname{editor}
}{}

\begin{document}

\title[ClinicalBERT]{ClinicalBERT: Modeling Clinical Notes \\ and Predicting Hospital Readmission}

\author{Kexin Huang}
\affiliation{
\institution{Health Data Science, Harvard T.H. Chan School of Public Health}
}

\author{Jaan Altosaar}
\affiliation{\institution{Department of Physics,\\ Princeton University}}

\author{Rajesh Ranganath}
\affiliation{\institution{Courant Institute of Mathematical Science, New York University}} 

\begin{abstract}

Clinical notes contain information about patients beyond structured data such as lab values or medications. However, clinical notes have been underused relative to structured data, because notes are high-dimensional and sparse. We aim to develop and evaluate a continuous representation of clinical notes. Given this representation, our goal is to predict 30-day hospital readmission at various timepoints of admission, including early stages and at discharge. We apply \gls{bert} to clinical text. Publicly-released \gls{bert} parameters are trained on standard corpora such as Wikipedia and BookCorpus, which differ from clinical text. We therefore pre-train \gls{bert} using clinical notes and fine-tune the network for the task of predicting hospital readmission. This defines ClinicalBERT. ClinicalBERT uncovers high-quality relationships between medical concepts, as judged by physicians.  ClinicalBERT outperforms various baselines on 30-day hospital readmission prediction using both discharge summaries and the first few days of notes in the intensive care unit on various clinically-motivated metrics. The attention weights of ClinicalBERT can also be used to interpret predictions. To facilitate research, we open-source model parameters, and scripts for training and evaluation. ClinicalBERT is a flexible framework to represent clinical notes. It improves on previous clinical text processing methods and with little engineering can be adapted to other clinical predictive tasks.
\end{abstract}

\maketitle

\section{Introduction}
An \gls{ehr} stores patient information; it can save money, time, and lives~\cite{Pedersen1336}. Data is added to an \gls{ehr} daily, so analyses may benefit from machine learning. Machine learning techniques leverage structured features in \gls{ehr} data, such as lab results or electrocardiography measurements, to uncover patterns and improve predictions~\cite{1706.03446,xiao2018opportunities,yu2018artificial}. However, unstructured, high-dimensional, and sparse information such as clinical notes are difficult to use in clinical machine learning models. Our goal is to create a framework for modeling clinical notes that can uncover clinical insights and make medical predictions.

Clinical notes contain significant clinical value~\cite{boag2018whats,weng2017medical,pmlr-v85-liu18b,WANG201812}. A patient might be associated with hundreds of notes within a stay and over their history of admissions. Compared to structured features, clinical notes provide a richer picture of the patient since they describe symptoms, reasons for diagnoses, radiology results, daily activities, and patient history. Consider clinicians working in the intensive care unit, who need to make decisions under time constraints. Making accurate clinical predictions may require reading a large volume of clinical notes. This can add to a doctor’s workload, so tools that make accurate predictions based on clinical notes might be useful in practice.

\begin{figure*}
    \centering
    \includegraphics[width=\linewidth]{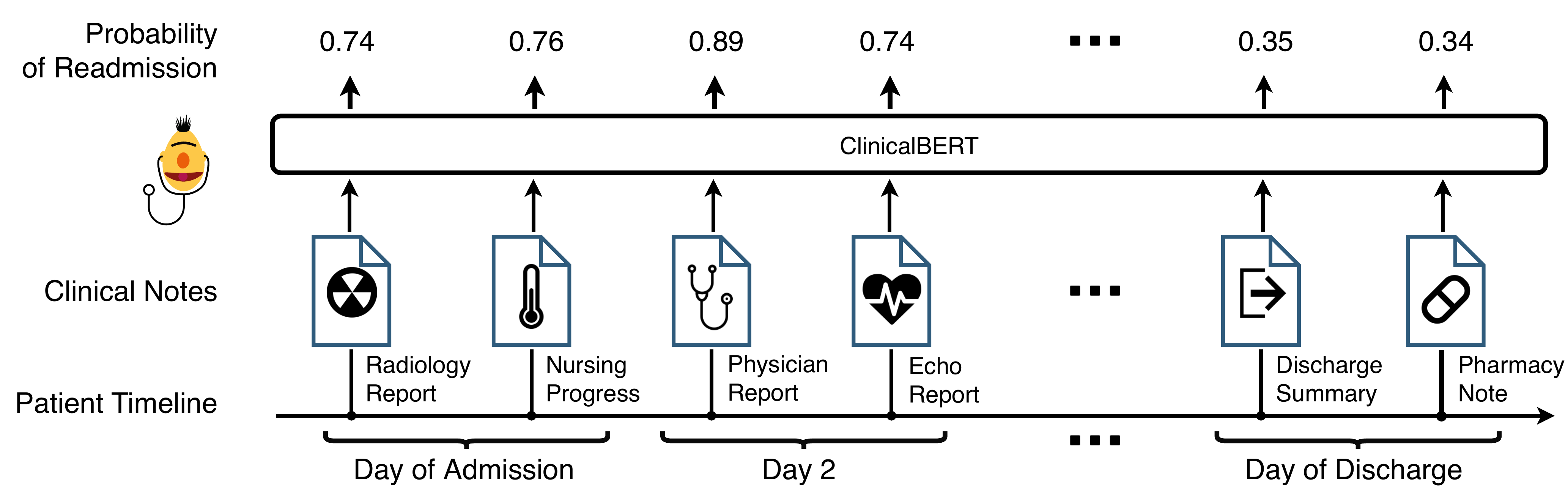}
    \caption{ClinicalBERT learns deep representations of clinical notes that are useful for tasks such as readmission prediction. In this example, care providers add notes to an electronic health record during a patient’s admission, and the model dynamically updates the patient’s risk of being readmitted within a 30-day window.}
    \label{fig:1}
\end{figure*}

Hospital readmission lowers patients’ quality of life and wastes money~\cite{anderson1984hospital,zuckerman2016readmissions}. One estimate puts the financial burden of readmission at \$17.9 billion and the fraction of avoidable admissions at 76\%~\cite{BasuRoy:2015:DHC:2783258.2788585}. Accurately predicting readmission has clinical significance, as it may improve efficiency and reduce the burden on intensive care unit doctors. We develop a discharge support model, ClinicalBERT, that processes patient notes and dynamically assigns a risk score of whether the patient will be readmitted within 30 days (\Cref{fig:1}). As physicians and nurses write notes about a patient, ClinicalBERT processes the notes and updates the risk score of readmission. This score can inform provider decisions, such as whether to intervene. Besides readmission, ClinicalBERT can be adapted to other tasks such as diagnosis prediction, mortality risk estimation, or length-of-stay assessment.
\subsection{Background}
Electronic health records are useful for risk prediction~\cite{Goldstein2017OpportunitiesAC}. Clinical notes in such electronic health records use abbreviations, jargon, and have an unusual grammatical structure. Building models that learn useful representations of clinical text is a challenge~\cite{10.1136/amiajnl-2011-000465}. Bag-of-words assumptions have been used to model clinical text~\cite{cite-key2}, in addition to log-bilinear word embedding models such as Word2Vec~\cite{1310.4546,inproceedings}. The latter word embedding models learn representations of clinical text using local contexts of words. But clinical notes are long and their words are interdependent~\cite{pmlr-v85-zhang18a}, so these methods cannot capture the long-range dependencies needed to capture clinical meaning.

Natural language processing methods where representations include global, long-range information can yield boosts in performance on clinical tasks~\cite{1802.05365,Radford2018ImprovingLU,1810.04805}. Modeling clinical notes requires capturing interactions between distant words. The need to model this long-range structure makes clinical notes suitable for contextual representations like bidirectional encoder representations from transformers (\gls{bert})~\cite{1810.04805}. \citet{1901.08746} apply \gls{bert} to biomedical literature, and~\cite{10.1093/jamia/ocz096} use \gls{bert} to enhance clinical concept extraction.
Concurrent to our work, \citet{1904.03323} also apply \gls{bert} to clinical notes; we evaluate and adapt ClinicalBERT to the clinical task of readmission and pre-train on longer sequence lengths.

Methods to evaluate models of clinical notes are also relevant to ClinicalBERT. \citet{WANG201812,W16-2922} evaluate the quality of biomedical embeddings by computing correlations between doctor-rated relationships and embedding similarity scores. We adopt similar evaluation techniques in our work.

Good representations of clinical text require good performance on downstream tasks. We use 30-day hospital readmission prediction as a case study since it is of clinical importance. We refer readers to \citet{FUTOMA2015229} for comparisons of traditional machine learning methods such as random forests and neural networks on hospital readmission tasks. Work in this area has focused on integrating a multitude of covariates about a patient into a model~\cite{10.1093/jamia/ocv110}. \citet{caruana2015intelligible} develop an interpretable model for readmission prediction based on generalized additive models and highlight the need for intelligible clinical predictions. \citet{cite-key1} predict readmission using a standard ontology from notes alongside structured information. Much of this previous work uses information at discharge, whereas ClinicalBERT can predict readmission during a patient’s stay.

\subsection{Significance}
ClinicalBERT improves readmission prediction over methods that center on discharge summaries. Making a prediction using a discharge summary at the end of a stay means that there are fewer opportunities to reduce the chance of readmission. To build a clinically-relevant model, we define a task of predicting readmission at any timepoint since a patient was admitted. To evaluate models on readmission prediction, we define a metric motivated by a clinical challenge. Medicine suffers from alarm fatigue~\cite{sendelbach2013alarm,10.1093/jamia/ocw150}. This means useful classification rules for medicine need to have high positive predictive value (precision). We evaluate model performance at a fixed positive predictive value. We show that ClinicalBERT has the highest recall compared to popular methods for representing clinical notes. ClinicalBERT can be readily applied to other tasks such as mortality prediction and disease prediction. In addition, ClinicalBERT attention weights can be visualized to understand which elements of clinical notes are relevant to a prediction.

ClinicalBERT is \gls{bert}~\cite{1810.04805} specialized to clinical notes. Clinical notes are lengthy and numerous, and the computationally-efficient architecture of \gls{bert} can model long-term dependencies. Compared to two popular models of clinical text, Word2Vec and FastText, ClinicalBERT more accurately captures clinical word similarity. We describe one way to scale up ClinicalBERT to handle large collections of clinical notes for clinical prediction tasks. In a case study of hospital readmission prediction, ClinicalBERT outperforms competitive deep language models. We open source ClinicalBERT\footnote{\url{https://github.com/kexinhuang12345/clinicalBERT}} pre-training and readmission model parameters along with scripts to reproduce results and apply the model to new tasks.

\section{Methods}
ClinicalBERT learns deep representations of clinical text. These representations can uncover clinical insights (such as predictions of disease), find relationships between treatments and outcomes, or create summaries of corpora. ClinicalBERT is an application of the \gls{bert} model~\cite{1810.04805} to clinical corpora to address the challenges of clinical text. Representations are learned using medical notes and further processed for clinical tasks; we demonstrate ClinicalBERT on the task of hospital readmission prediction.

\subsection{BERT Model}
\gls{bert} is a deep neural network that uses the transformer encoder architecture~\cite{1706.03762} to learn embeddings for text. We omit a detailed description of the architecture; it is described in~\cite{1706.03762}. The transformer encoder architecture is based on a self-attention mechanism. The pre-training objective function for the model is defined by two unsupervised tasks: masked language modeling and next sentence prediction. The text embeddings and model parameters are fit using stochastic optimization. For downstream tasks, the fine-tuning phase is problem-specific; we describe a fine-tuning task specific to clinical text.

\subsection{Clinical Text Embedding}
A clinical note input to ClinicalBERT is represented as a collection of tokens. These tokens are subword units extracted from text in a preprocessing step~\cite{sentence}. In ClinicalBERT, a token in a clinical note is represented as a sum of the token embedding, a learned segment embedding, and a position embedding. When multiple sequences of tokens are fed to ClinicalBERT, the segment embedding identifies which sequence a token is associated with. The position embedding of a token is a learned set of parameters corresponding to the token’s position in the input sequence (position embeddings are shared across tokens). A classification token [CLS] is inserted in front of every sequence of input tokens for use in classification tasks.

\subsection{Self-Attention Mechanism}
The attention function is computed on an input sequence using the embeddings associated with the input tokens. The attention function takes as input a set of queries, keys, and values. To construct the queries, keys, and values, every input embedding is multiplied by learned sets of weights (it is called ‘self’ attention because the values are the same as the keys and queries). For a single query, the output of the attention function is a weighted combination of values. The query and a key determine the weight for a value. Denote a set of queries, keys, and values by Q, K, and V. The attention function is
\begin{equation}
    \mathrm{Attention}(Q,K,V) = \mathrm{softmax}(\frac{QK^T}{\sqrt{d}} V),
\end{equation}
where d is the dimensionality of the queries, keys, and values. This function can be computed efficiently and can capture long-range interactions between any two elements of the input sequence~\cite{1706.03762}. The length and complex patterns in clinical notes makes the transformer architecture with self-attention a good choice. (We later describe how this attention mechanism can allow interpretation of ClinicalBERT predictions.)

\begin{figure}
    \centering
    \includegraphics[width = \linewidth]{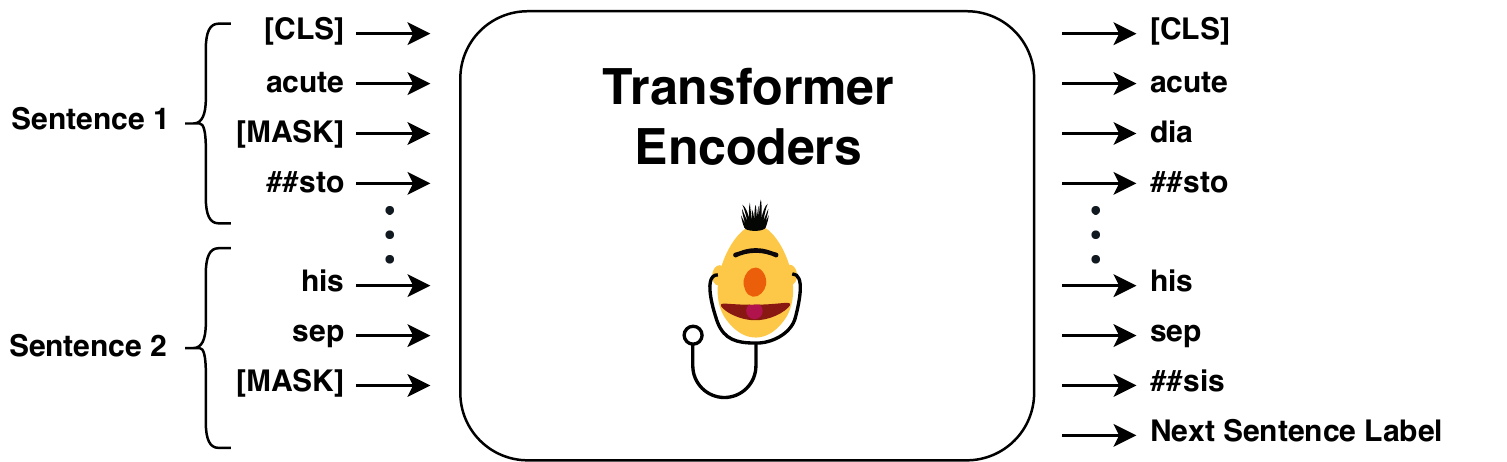}
    \caption{ClinicalBERT learns deep representations of clinical text using two unsupervised language modeling tasks: masked language modeling and next sentence prediction. In masked language modeling, a fraction of input tokens are held out for prediction; in next sentence prediction, ClinicalBERT predicts whether two input sentences are consecutive.}
    \label{fig:2}
\end{figure}
\subsection{Pre-training ClinicalBERT}
The quality of learned representations of text depends on the text the model was trained on. \gls{bert} is trained on BooksCorpus and Wikipedia. But these datasets are distinct from clinical notes, as jargon and abbreviations prevail: clinical notes have different syntax and grammar than books or encyclopedias. These differences make clinical notes hard to understand without expertise. ClinicalBERT is pre-trained on clinical notes as follows.

ClinicalBERT uses the same pre-training tasks as~\cite{1810.04805}. Masked language modeling means masking some input tokens and training the model to predict the masked tokens. In next sentence prediction, two sentences are fed to the model. The model predicts whether these sentences are consecutive. The pre-training objective function is the sum of the log-likelihood of the predicted masked tokens and the log-likelihood of the binary variable indicating whether two sentences are consecutive.
\begin{figure}
    \centering
    \includegraphics[width=\linewidth]{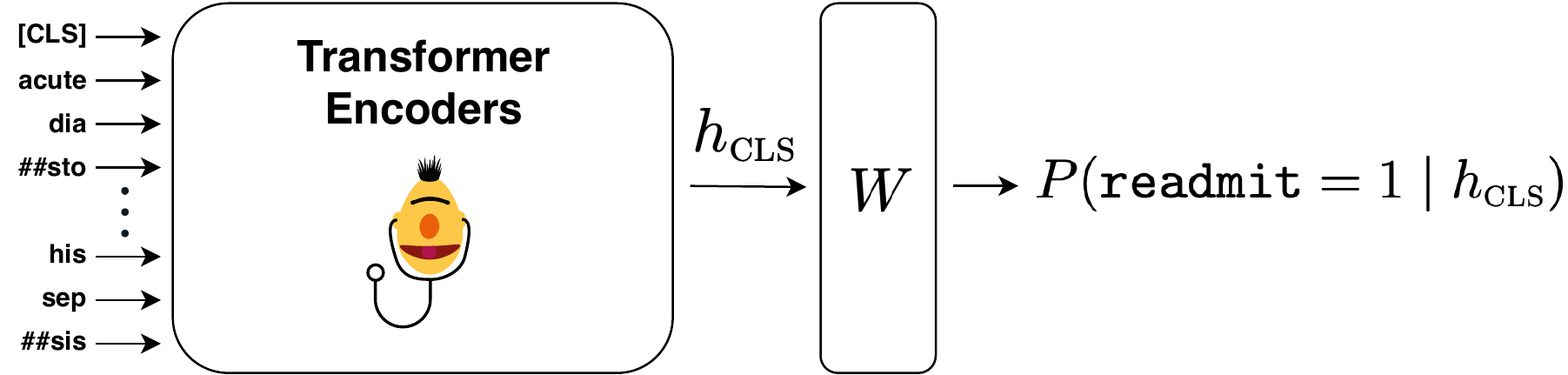}
    \caption{ClinicalBERT models clinical notes and can be readily adapted to clinical tasks such as predicting 30-day readmission. The model is fed a patient’s clinical notes, and the patient’s risk of readmission within a 30-day window is predicted using a linear layer applied to the classification representation $h_\mathrm{[CLS]}$ learned by ClinicalBERT. This fine-tuning task is described in \Cref{eq:finetune}}
    \label{fig:3}
\end{figure}

\subsection{Fine-tuning ClinicalBERT}
After pre-training, ClinicalBERT is fine-tuned on a clinical task: readmission prediction. Let readmit be a binary indicator of readmission of a patient in the next 30 days. Given clinical notes as input, the output of ClinicalBERT is used to predict the probability of readmission:
\begin{equation}\label{eq:finetune}
    P(\texttt{readmit} = 1 | h_\mathrm{[CLS]}) = \sigma(Wh_\mathrm{[CLS]})
\end{equation}
where $\sigma$ is the sigmoid function, $h_\mathrm{[CLS]}$ is the output of the model corresponding to the classification token, and W  is a parameter matrix. The model parameters are fine-tuned to maximize the log-likelihood of this binary classifier.

\section{Empirical Study}
\subsection{Data}
We use the \gls{mimic} dataset~\cite{cite-key_mimic}. \gls{mimic} consists of the electronic health records of 58,976 unique hospital admissions from 38,597 patients in the intensive care unit of the Beth Israel Deaconess Medical Center between 2001 and 2012. There are 2,083,180 de-identified notes associated with the admissions. Preprocessing of the clinical notes is described in S2. If text that exists in the test set of the fine-tuning task is used for pre-training, then training and test metrics will not be independent. To avoid this, admissions are split into five folds for independent runs, with four folds for pre-training (and training during fine-tuning) and the fifth for testing during fine-tuning.

\subsection{Empirical Study I: Language Modeling and Clinical Word Similarity}

We developed ClinicalBERT, a model of clinical notes whose representations can be used for clinical tasks. Before evaluating its performance as a model of readmission, we study its performance in two experiments. First, we find that ClinicalBERT outperforms \gls{bert} in clinical language modeling. Then we compare ClinicalBERT to popular word embedding models using a clinical word similarity task. The relationships between medical concepts learned by ClinicalBERT correlate with human evaluations of similarity.

\begin{table}[t]
    \centering
    \caption{ClinicalBERT improves over \gls{bert} on clinical language modeling. We report the five-fold average accuracy of masked language modeling (predicting held-out tokens) and next sentence prediction (a binary prediction of whether two sentences are consecutive), on the \gls{mimic} corpus of clinical notes. }
    \begin{tabular}{l|cc}
    \toprule
    Model & Language modeling & Next sentence prediction  \\ \hline
    ClinicalBERT & 0.857 $\pm$ 0.002 & 0.994 $\pm$ 0.003 \\
    \gls{bert} & 0.495 $\pm$ 0.007 & 0.539  $\pm$ 0.006 \\
    \bottomrule
    \end{tabular}
    \label{tab:1}
\end{table}
\subsubsection{Clinical Language Modeling.}
We report the five-fold average accuracy of the masked language modeling and next sentence prediction tasks on the \gls{mimic} data in \Cref{tab:1}. \gls{bert} underperforms, as it was not trained on clinical text, highlighting the need for building models tailored to clinical data such as ClinicalBERT.

\subsubsection{Qualitative Analysis.} 
\begin{figure}
    \centering
    \includegraphics[width = \linewidth]{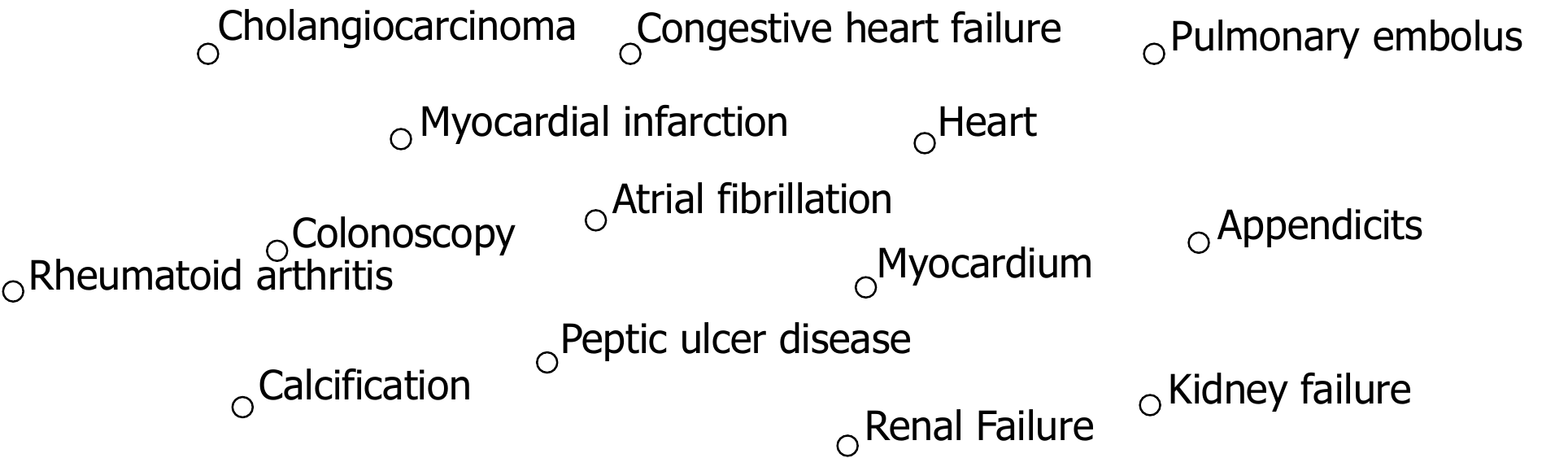}
    \caption{ClinicalBERT reveals interpretable patterns in medical concepts. The model is trained on clinical notes from \gls{mimic}, and the embeddings of clinical terms from the dataset in~\cite{PEDERSEN2007288} are plotted using the t-distributed stochastic neighbor embedding algorithm for dimensionality reduction~\cite{article_tsne}. We highlight a subset of the plot centered on a cluster of terms relating to heart conditions such as myocardial infarction, heart failure, and kidney failure.}
    \label{fig:4}
\end{figure}

We test ClinicalBERT on data collected to assess medical term similarity~\cite{PEDERSEN2007288}. The data is 30 pairs of medical terms whose similarity is rated by physicians. To compute an embedding for a medical term, ClinicalBERT is fed a sequence of tokens corresponding to the term. Following~\cite{1810.04805}, the sum of the last four hidden states of ClinicalBERT encoders is used to represent each medical term. Medical terms vary in length, so the average is computed over the hidden states of subword units. This results in a fixed 768-dimensional vector for each medical term. We visualize the similarity of medical terms using dimensionality reduction~\cite{article_tsne}, and display a cluster heart-related concepts in \Cref{fig:4}. Heart-related concepts such as myocardial infarction, atrial fibrillation, and myocardium are close together; renal failure and kidney failure are also close. This demonstrates that ClinicalBERT captures some clinical semantics.

\subsubsection{Quantitative Analysis.} We benchmark embedding models using the clinical concept dataset in~\cite{PEDERSEN2007288}. The data consists of concept pairs, and the similarity of a pair is rated by physicians, with a score ranging from 1.0 to 4.0 (least similar to most similar). To evaluate representations of clinical text, we calculate the similarity between two concepts’ embeddings a and b using cosine similarity,
\begin{equation}
    \mathrm{Similarity}(a,b) = \frac{a\cdot b}{\Vert a \Vert \Vert b \Vert}
\end{equation}
We calculate the Pearson correlation between physician ratings of medical concept similarity and the cosine similarity between model embeddings. Models with high correlation capture human-rated similarity between clinical terms. \citet{WANG201812} conducts a similar evaluation on this data using Word2Vec word embeddings~\cite{1310.4546} trained on clinical notes, biomedical literature, and Google News. However, this work relies on a private clinical note dataset from The Mayo Clinic to train the Word2Vec model. For a fair comparison with ClinicalBERT, we retrain the Word2Vec model using clinical notes from \gls{mimic}. The Word2Vec model is trained on 2.8B words from \gls{mimic} with the same hyperparameters as~\cite{WANG201812}. Word2Vec cannot handle out-of-vocabulary words; we ignore the three medical pairs in the clinical concepts dataset that do not have embeddings (correlation is computed using the remaining 27 medical pairs). Because of this shortcoming, we also train a FastText model~\cite{DBLP:journals/corr/BojanowskiGJM16} on \gls{mimic}, which models out-of-vocabulary words using subword units. FastText and Word2Vec are trained on the full \gls{mimic} data, so we also pre-train ClinicalBERT on the full data for comparison. \Cref{tab:2} shows how these models correlate with physician, with ClinicalBERT more accurately correlating with physician judgment.

\begin{table}[t]
    \centering
     \caption{ClinicalBERT captures physician-assessed relationships between clinical terms. The Pearson correlation is computed between the cosine similarity of embeddings learned by models of clinical text and physician ratings of the similarity of medical concepts in the dataset of~\cite{PEDERSEN2007288}. These numbers are comparable to the best result, 0.632, from~\cite{WANG201812}.
}
    \begin{tabular}{l|c}
    \toprule
    Model & Pearson correlation \\ \hline
    ClinicalBERT & 0.670 \\
    Word2Vec & 0.553 \\
    FastText & 0.487 \\
    \bottomrule
    \end{tabular}
   
    \label{tab:2}
\end{table}

\subsection{Empirical Study II: 30-Day Hospital Readmission Prediction}
The representations learned by ClinicalBERT can help address problems in the clinic. We build a model to predict hospital readmission from clinical notes. Compared to benchmark language models, ClinicalBERT accurately predicts readmission. Further, ClinicalBERT predictions can be interrogated by visualizing attention weights to reveal interpretable patterns in medical data.

\subsubsection{Cohort.} We select a patient cohort from \gls{mimic} using patient covariates. The binary readmit label associated with each patient admission is computed as follows. Admissions where a patient is readmitted within 30 days are labeled $\texttt{readmit}=1$. All other patient admissions are labeled zero, including patients with appointments within 30 days (to model unexpected readmission). In-hospital death precludes readmission, so admissions with deaths are removed. Newborn patients account for 7,863 admissions. Newborns are in the neonatal intensive care unit, where most undergo testing and are sent back for routine care. This leads to a different distribution of clinical notes and readmission labels; we filter out newborns and focus on non-newborn readmissions. The final cohort contains 34,560 patients with 2,963 positive readmission labels and 42,358 negative labels.

\subsubsection{Scalable Readmission Prediction.} Patients are often associated with many notes. ClinicalBERT has a fixed length of input sequence, so notes are concatenated and split to this maximum length. Predictions for patients with many notes are computed by binning the predictions on each subsequence. The probability of readmission for a patient is computed as follows. For a patient whose notes are split into n subsequences, ClinicalBERT outputs a probability for each subsequence. The probability of readmission is computed using the predictions for each subsequence:
\begin{equation}\label{eq:predict-readmit}
   P(\texttt{readmit} = 1 \mid h_\text{patient}) = \frac{P_{\text{max}}^n+P_{\text{mean}}^n n/c}{1+n/c} \, ,
\end{equation}
The scaling factor $c$ controls the influence of the number of subsequences n, and $ h_\text{patient})$ is the implicit ClinicalBERT representation of all of a patient’s notes. The maximum and mean probabilities of readmission over n subsequences are $P_\text{max}^n$ and $P_\text{mean}^n$.

Computing readmission probability using \Cref{eq:predict-readmit} outperforms predictions using the mean for each subsequence by 3–8\%. This formula is motivated by observations: some subsequences do not contain information about readmission (such as tokens corresponding to progress reports), whereas others do. The risk of readmission should be computed using subsequences that correlate with readmission, and the effect of unimportant subsequences should be minimized. This is accomplished by using the maximum probability over subsequences. Second, noise in subsequences decreases performance. For example, consider the case where one noisy subsequence has a prediction of 0.8, but all other subsequences have predictions close to zero. Using only the maximum would lead to a false prediction if the maximum is due to noise, so we include the average probability of readmission across subsequences. This leads to a trade-off between the mean and maximum probabilities of readmission in \Cref{eq:predict-readmit}. Finally, if there are a large number of subsequences (for a patient with many clinical notes), there is a higher probability of a noisy maximum probability of readmission. This means longer sequences may need a larger weight on the mean prediction. We include this weight as an   $n/c$  scaling factor, with c accounting for patients with many notes. The denominator results from normalizing the risk score to the unit interval. The parameter c is selected using the validation set; $c=2$ was selected.

\subsubsection{Evaluation.} For validation and testing, the cohort is split into five folds. In each fold 20\% is used for validation (10\%) and test (10\%) sets, with the rest for training. Each model is evaluated using three metrics:

\begin{enumerate}
    \item Area under the receiver operating characteristic curve (AUROC): the area under the true positive rate versus the false positive rate.
    \item Area under the precision-recall curve (AUPRC): the area under the plot of precision versus recall.
    \item Recall at precision of 80\% (RP80): for readmission prediction, false positives are important. To minimize the number of false positives and hence minimize the risk of alarm fatigue, we fix precision to 80\% (or, 20\% false positives in the positive class predictions). This threshold is used to calculate recall. This leads to a clinically-relevant metric that enables building models that minimize the false positive rate.
\end{enumerate}

\subsubsection{Models.} We compare ClinicalBERT to three competitive models. \citet{boag2018whats} conclude that a bag-of-words model and a \gls{lstm} model with Word2Vec embeddings work well for predictive tasks on \gls{mimic} clinical notes. We also compare to \gls{bert} with trainable weights. Training details are in \Cref{appendix:1}.
\begin{enumerate}
    \item ClinicalBERT: the model parameters include the weights of the encoder network and the learned classifier weights.
    \item Bag-of-words: this method uses word counts to represent a note. The 5,000 most frequent words are used as features. Logistic regression with L2 regularization is used to predict readmission.
    \item \Gls{bi-lstm} and Word2Vec ~\cite{Schuster1997BidirectionalRN,doi:10.1162/neco.1997.9.8.1735}: a \gls{bi-lstm} is used to model words in a sequence. The final hidden layer is used to predict readmission.
    \item \gls{bert}: this is what ClinicalBERT is based on, but \gls{bert} is pre-trained not on clinical notes but standard language corpora.
\end{enumerate}
We also compared to ELMo~\cite{1802.05365}, where a standard 1,024-dimensional embedding for each text subsequence is computed and a neural network classifier is used to fit the training readmission labels. The performance was much worse, and we omit these results. This may be because the weights in ELMo are not learned, and the fixed-length embedding may not be able to store the information needed for a classifier to detect signal from long and complex clinical text.

\begin{table}[t]
    \centering
    \caption{ClinicalBERT accurately predicts 30-day readmission using discharge summaries. The mean and standard deviation of 5-fold cross validation is reported. ClinicalBERT outperforms the bag-of-words model, the \gls{bi-lstm}, and \gls{bert} deep language models.
}
    \begin{tabular}{l|ccc}
    \toprule
    Model & AUROC & AUPRC & RP80 \\ \hline
    ClinicalBERT & 0.714 $\pm$ 0.018 & 0.701 $\pm$ 0.021 &	0.242 $\pm$ 0.111  \\
    Bag-of-words & 0.684 $\pm$ 0.025 & 0.674 $\pm$ 0.027 & 0.217 $\pm$ 0.119 \\
    \gls{bi-lstm}	& 0.694 $\pm$ 0.025	& 0.686 $\pm$ 0.029 &	0.223 $\pm$ 0.103 \\
    \gls{bert} & 0.692 $\pm$ 0.019 & 0.678 $\pm$ 0.016 & 0.172 $\pm$ 0.101 \\
    \bottomrule
    \end{tabular}
    \label{tab:3}
\end{table}

\subsubsection{Readmission Prediction with Discharge Summaries.} Discharge summaries contain essential information of patient admissions since they are used by the post-hospital care team and by doctors in future visits~\cite{van2002effect}. The summary may contain information like a patient’s discharge condition, procedures, treatments, and significant findings~\cite{kind2008documentation}. This means discharge summaries should have predictive value for hospital readmission.  \Cref{tab:3} shows that ClinicalBERT outperforms competitors in terms of precision and recall on a task of readmission prediction using patient discharge summaries.

\begin{table*}[]
    \centering
    \caption{ClinicalBERT outperforms competitive baselines on readmission prediction using clinical notes from early on within patient admissions. In \gls{mimic} data, admission and discharge times are available, but clinical notes do not have timestamps. The cutoff time indicates the range of admission durations that are fed to the model from early in a patient’s admission. For example, in the 24–48h column, the model may only take as input a patient’s notes up to 36h because of that patient’s specific admission time. Metrics are reported as the mean and standard deviation of 5 independent runs.
}
\begin{tabular}{l|cccc}
    \toprule
    Model & Cutoff time & AUROC	& AUPRC & RP80 \\ \hline
\multirow{2}{*}{ClinicalBERT} & 24–48h & 0.674 $\pm$ 0.038 & 0.674 $\pm$ 0.039 & 0.154 $\pm$ 0.099 \\
& 48–72h & 0.672 $\pm$ 0.039&0.677 $\pm$ 0.036 & 0.170 $\pm$ 0.114 \\ \hline
\multirow{2}{*}{Bag-of-words} & 24–48h  & 0.648 $\pm$ 0.029 &	0.650 $\pm$ 0.027 & 0.144 $\pm$ 0.094 \\
 & 48–72h & 0.654 $\pm$ 0.035 & 0.657 $\pm$ 0.026 & 0.122 $\pm$ 0.106 \\ \hline
\multirow{2}{*}{\gls{bi-lstm}} & 24–48h  & 0.649 $\pm$ 0.044	& 0.660 $\pm$ 0.036 &	0.143 $\pm$ 0.080 \\
 & 48–72h	& 0.656 $\pm$ 0.035	 & 0.668 $\pm$ 0.028	& 0.150 $\pm$ 0.081 \\ \hline
\multirow{2}{*}{\gls{bert}} & 24–48h & 0.659 $\pm$ 0.034	& 0.656 $\pm$ 0.021&	0.141 $\pm$ 0.080 \\
&	48–72h &	0.661 $\pm$ 0.028&	0.668 $\pm$ 0.021	& 0.167 $\pm$ 0.088 \\
\bottomrule
\end{tabular}

\label{tab:4}
\end{table*}

\subsubsection{Readmission Prediction with Early Clinical Notes.} Discharge summaries can be used to predict readmission, but may be written after a patient has left the hospital. Therefore, discharge summaries are not useful for intervention—doctors cannot intervene when a patient has left the hospital. Models that dynamically predict readmission in the early stages of a patient’s admission are relevant to clinicians. For the second set of readmission prediction experiments, a maximum of the first 48 or 72 hours of a patient’s notes are concatenated. These concatenated notes are used to predict readmission. Since we separate notes into subsequences of the same length, the training set consists of all subsequences up to a cutoff time. The model is tested given notes up to 24–48h or 48–72h of a patient’s admission. We do not consider 0-24h cutoff time because there may be too few notes for good predictions. Note that readmission predictions from a model are not actionable if a patient has been discharged. For evaluation, patients that are discharged within the cutoff time are filtered out.
Models of readmission prediction are evaluated using the metrics. \Cref{tab:4} shows that ClinicalBERT outperforms competitors in both experiments. The AUROC and AUPRC results show that ClinicalBERT has more confidence and higher accuracy. At a fixed rate of false alarms, ClinicalBERT recalls more patients that have been readmitted, and its performance increases as the length of admissions increases and the model has access to more clinical notes.

\begin{figure}
    \centering
    \includegraphics[width = \linewidth]{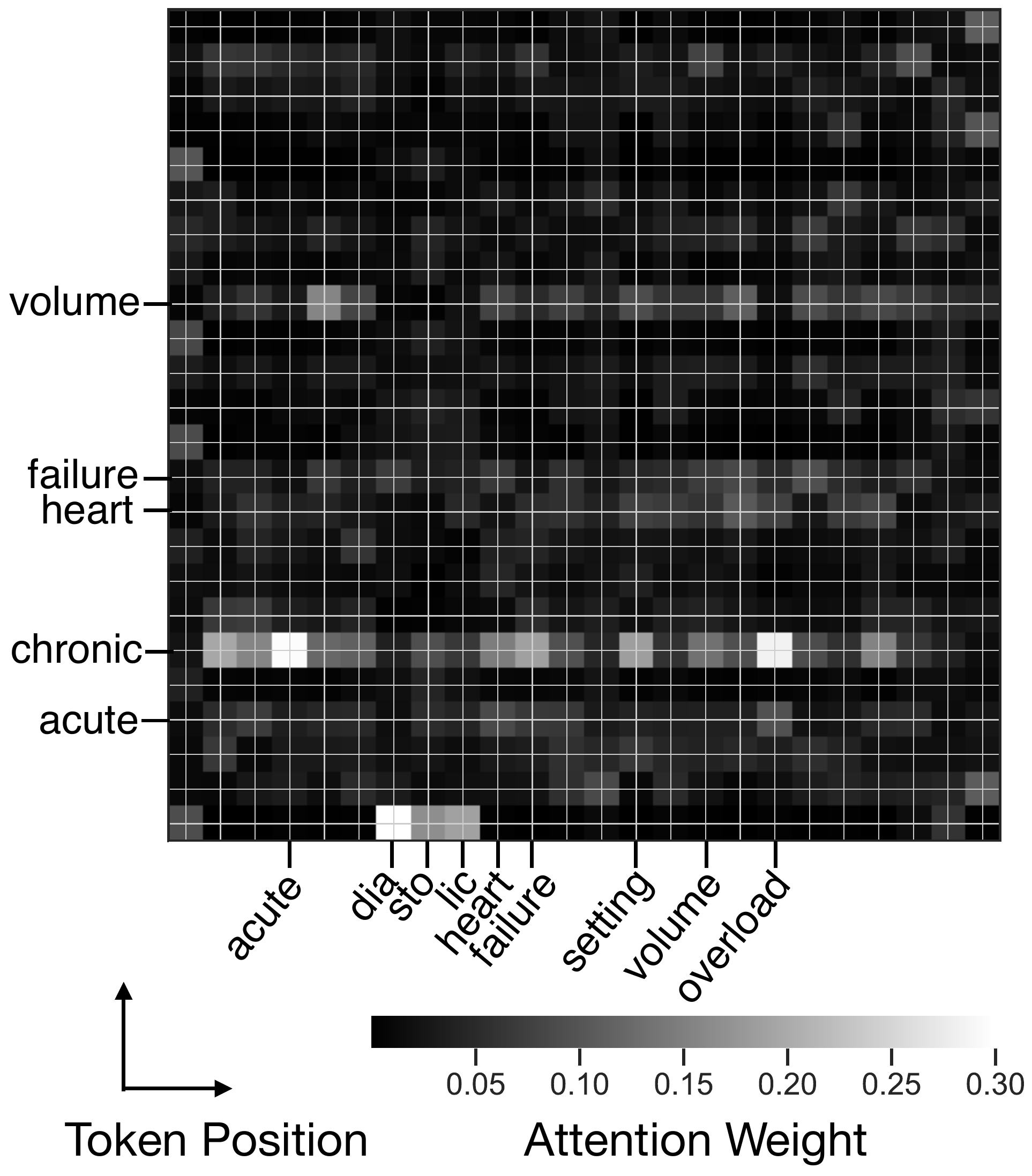}
    \caption{ClinicalBERT provides interpretable predictions, by revealing which terms in clinical notes are predictive of patient readmission. The self-attention mechanisms in ClinicalBERT can be used to interpret model predictions on clinical notes. The input sentence “he has experienced acute chronic diastolic heart failure in the setting of volume overload due to his sepsis.” is fed to the model (this sentence is representative of a clinical note found in \gls{mimic}). \Cref{eq:attention-weight} is used to compute a distribution over tokens in this sentence, where every query token is itself a token in the same input sentence. In the panel, we show one of the self-attention mechanisms in ClinicalBERT, and only label terms that have high attention weight. The x-axis labels are query tokens and the y-axis labels are key tokens. }
    \label{fig:5}
\end{figure}

\subsubsection{Interpretability.} Clinician mistrust of data-driven methods is sensible: predictions from a neural network are difficult to understand for humans, and it is not clear why a model makes a certain prediction or what parts of the data are most informative. ClinicalBERT uses several attention mechanisms which can be used to inspect predictions by visualizing terms correlated with hospital readmission. For a clinical note fed to ClinicalBERT, attention mechanisms compute a distribution over every term in a sentence, given a query term. For a given query vector q computed from an input token, the attention weight distribution is defined as
\begin{equation}\label{eq:attention-weight}
    \text{AttentionWeight}(q,K)=\textrm{softmax}\bigg(\frac{qK^\top}{\sqrt{d}}\bigg).
\end{equation}
The attention weights are used to compute the weighted sum of values. A high attention weight between a query and key token means the interaction between these tokens is predictive of readmission. In the ClinicalBERT encoder, there are 144 self-attention mechanisms (or, 12 multi-head attention mechanisms for each of the 12 transformer encoders). After training, each mechanism specializes to different patterns in clinical notes that are indicative of readmission.

To illustrate, a sentence representative of a \gls{mimic} note is fed to ClinicalBERT. Both the queries and keys are the tokens in the sentence. Attention weight distributions for every query are computed using \Cref{eq:attention-weight} and visualized in \Cref{fig:5}. The panel shows an attention mechanism that is activated for the word ‘chronic’ and ‘acute’ given any query term. This means some attention heads focus on for specific predictive terms, a similar computation to a bag-of-words model. Intuitively, the word ‘chronic’ is a predictor of readmission.

\section{Guidelines on using ClinicalBERT in Practice}
ClinicalBERT is pre-trained on \gls{mimic}, which consists of patients from ICUs in one Boston hospital. As notes vary by institution and clinical setting (e.g. ICU vs outpatient), to use ClinicalBERT in practice we recommend training ClinicalBERT using the private \gls{ehr} dataset available at the practitioner’s institution. After fitting the model, ClinicalBERT can be used for downstream clinical tasks (e.g. mortality prediction or length-of-stay prediction). We include a tutorial for adapting ClinicalBERT for such downstream classification tasks in the repository.

\section{Discussion}
We developed ClinicalBERT, a model for learning deep representations of clinical text. Empirically, ClinicalBERT is an accurate language model and captures physician-assessed semantic relationships in clinical text. In a 30-day hospital readmission prediction task, ClinicalBERT outperforms a deep language model and yields a large relative increase on recall at a fixed rate of false alarms. Future work includes engineering to scale ClinicalBERT to capture dependencies in long clinical notes; the max and sum operations in \Cref{eq:predict-readmit} may not capture correlations within long notes. Finally, note that the \gls{mimic} dataset we use is small compared to the large volume of clinical notes available internally at hospitals. Rather than using pre-trained \gls{mimic} ClinicalBERT embeddings, this suggests that the use of ClinicalBERT in hospitals should entail re-training the model on this larger collection of notes for better performance. The publicly-available ClinicalBERT model parameters can be used to evaluate performance on clinically-relevant prediction tasks based on clinical notes.

\section{Acknowledgements}
We thank Noémie Elhadad for helpful discussion. Grass icon by Milinda Courey from the Noun Project.

\printbibliography

\appendix

\section{Hyperparameters and training details}\label{appendix:1}
The parameters are initialized to the \gls{bert} Base parameters released by~\cite{1810.04805}; we follow their recommended hyper-parameter settings. The model dimensionality is 768. We use the Adam optimizer with a learning rate of $2 x 10^-5$. The maximum sequence length supported by the model is set to 512, and the model is first trained using shorter sequences. The details of constructing a sequence are in~\cite{1810.04805}. For efficient mini-batching that avoids padding mini-batch elements of variable lengths with too many zeros, a corpus is split into multiple sequences of equal lengths. Many sentences are packed into a sequence until the maximum length is reached; a sequence may be composed of many sentences. The next sentence prediction task defined in~\cite{1810.04805} might more accurately be termed a next sequence prediction task. Our ClinicalBERT model is first trained using a maximum sequence length of 128 for 100,000 iterations on the masked language modeling and next sentence prediction tasks, with a batch size 64. Next, the model is trained on longer sequences of maximum length 512 for an additional 100,000 steps with a batch size of 8. When using text that exists in the test set of the fine-tuning task for pre-training, the training and test set during fine-tuning will not be independent. To avoid this, admissions are split into five folds for independent runs, with four folds for pre-training and training during fine-tuning and the fifth for testing during fine-tuning. Hence, for each independent run, during pre-training, we use all the discharge summaries associated with admissions in the four folds. During fine-tuning for readmission task, ClinicalBERT is trained for three epochs with batch size 56 and learning rate $2 x 10^-5$. The binary classifier is a three layers neural network of shape 768 x 2048, 2048 x 768, and 768 x 1. We fine-tune ClinicalBERT with three epochs and early stopped on validation loss as the criteria.

For Bi-\gls{lstm}, for the input word embedding, the Word2Vec model is used. The Bi-\gls{lstm} has 200 output units, with a dropout rate of 0.1. The hidden state is fed into a global max pooling layer and a fully-connected layer with a dimensionality of 50, followed by a rectifier activation function. The rectifier is followed by a fully-connected layer with a single output unit with sigmoid activation function. The binary classification objective function is optimized using the Adam adaptive learning rate (40). The Bi-\gls{lstm} is trained for three epochs with a batch size of 64 with early stopping based on the validation loss.

For the empirical study, we use a server with 2 Intel Xeon E5-2670v2 2.5GHZ CPUs, 128GB RAM and 2 NVIDIA Tesla P40 GPUs.

\section{Preprocessing Notes for Pretraining ClinicalBERT}

ClinicalBERT requires minimal preprocessing. First, words are converted to lowercase and line breaks and carriage returns are removed. Then de-identified brackets and remove special characters like ==, -- are removed. The next sentence prediction pretraining task described in Section 5 requires two sentences at every iteration. The SpaCy sentence segmentation package is used to segment each note. Since clinical notes don’t follow rigid standard language grammar, we find rule-based segmentation has better results than dependency parsing-based segmentation. Various segmentation signs that misguide rule-based segmentators are removed (such as 1.2.) or replaced (M.D., dr. with MD, Dr). Clinical notes can include various lab results and medications that also contain numerous rule-based separators, such as 20mg, p.o., q.d.. To address this, segmentations that have less than 20 words are fused into the previous segmentation so that they are not singled out as different sentences.

\end{document}